\documentclass{isprs} 
\usepackage{subfigure}
\usepackage{setspace}
\usepackage{geometry} 
\usepackage{epstopdf}
\usepackage[labelsep=period]{caption}  
\usepackage[british]{babel} 
\usepackage[hang]{footmisc}
\usepackage{hyperref}

\usepackage{ifthen}
\newboolean{anonymizeIt} 
\setboolean{anonymizeIt}{false} 
\usepackage{wrapfig}

\begin{document}
\title{Seamless Indoor-Outdoor Mapping for INGENIOUS First Responders}

\date{}


\ifthenelse{\boolean{anonymizeIt}}
{
	\author{***** (for review, names must be rendered anonymous)}
	\address{**** (for review, affiliations must be rendered anonymous)}
}
{
	\author{
		Jürgen Wohlfeil, 
		Henry Meißner,
		Adrian Schischmanow,
		Thomas Kraft,
		Dirk Baumbach,
		Ines Ernst,
		Dennis Dahlke
	}

	\address{
		Institute of Optical Systems (OS), German Aerospace Center (DLR) Berlin, Germany \\ (juergen.wohlfeil, henry.meissner, adrian.schischmanow, thomas.kraft, dirk.baumbach, ines.ernst, dennis.dahlke)@dlr.de
	}

}


\abstract{

In several applications it is desired to have 3D models not only from the outdoor spaces but also from inside the building. In the context of First Responder enhancement in large scale natural and man-made disasters, a method is presented to achieve this goal with a high degree of automation. Therefore an autonomously flying aerial mapping system is combined with a person-carried indoor positioning system. Automatically recognized markers (AprilTags) are geo-referenced by the aerial system and their coordinates are sent to the ground-based system. By looking at the AprilTags before entering the building, the ground-based system is registered to world coordinates. Without the further need of any global positioning, it creates a point cloud from the indoor spaces that fits with the point could from the aerial view. This allows a co-visualization of both point-clouds as a seamless indoor-outdoor 3D model in real time.
}


\maketitle

\section{Introduction}\label{sec:introduction}

First Responders are often faced to challenging tasks in man made and natural disasters (e.g. earthquake, large scale industrial accident, terror attack, etc.). 
The EU Project “INGENIOUS” (The First Responder of the Future: a Next Generation Integrated Toolkit for Collaborative Response, increasing protection and augmenting operational capacity, \cite{INGENIOUS_website} aims helping First Responders being more effective and save more lives in such circumstances. At the same time, the First Responder's safety shall be improved. As the name suggests, many different technologies are developed in the project and fused to one integrated toolset.

Two of these technologies contributed by the DLR (German Aerospace Center) are presented in this paper with the focus on creation of seamless indoor-outdoor 3D models of the typical scenarios chosen for INGENIOUS:
\begin{enumerate}
	\item For large scale mapping MACS-SaR (the \underline{M}odular \underline{A}irborne \underline{C}amera \underline{S}ystem for \underline{S}earch \underline{a}nd \underline{R}escue) \cite{dlr111102} is used to generate a 3D model of the outdoor spaces. It also provides a basis to co-register outdoor and indoor point clouds.
	\item For seamless indoor and outdoor navigation IPS (\underline{I}ntegrated \underline{P}ositioning \underline{S}ystem) is used. It is a person-carried opto-inertial navigation system that can navigate without GNSS (Global Navigation Sattelite System) or infrastructure-based positioning sources. It also provides a point cloud of the travelled areas. \cite{dlr63430,dlr111874}.
\end{enumerate}

Both technologies aim to enhance the operational effectiveness and safety of First Responders by providing tools for seamless indoor-outdoor navigation and accurate situational awareness. This enables quicker decision-making and better coordination during disaster response operations.

\textbf{MACS-SaR in the context of INGENIOUS:}  MACS is a drone equipped with a high resolution camera system, an L2-GNSS receiver and an industrial grade IMU-system. This hard- and software setup delivers precise geo-referenced 3D point clouds for search and rescue parties directly after landing and can be visualized on-site or in the base of operation. Moreover, MACS-SaR automatically triangulates all images showing signalized AprilTags (see Section \ref{sec:covis}) and thusly establishing the base coordinate system for co-registration and co-visualization of IPS and MACS.

\textbf{IPS in the context of INGENIOUS:} The benefit of IPS in the context of this project is that the position (and orientation) of a First Responder carrying IPS is transmitted and visualized in the established Center of Operation and can also be provided and visualized to other team members. The First Responder carrying the IPS device can use it to take photos of, for example, discovered victims or structurally critical areas of the building. These photos are transmitted in real time to the Center of Operation together with the position and orientation they were taken. Finally also a sparse point cloud is provided by IPS. 
With further processing a dense point cloud can be created using Semi Global Matching (SGM) \cite{dlr22952}. 
By transmitting photos and point clouds in real time, IPS allows responders to share critical information about victims or hazards with team members and command centers, improving coordination and decision-making.
IPS is useful as it provides seamless indoor-outdoor navigation and can operate independently of global positioning (e.g., GPS, WiFi, etc.), which may be unavailable indoors during disasters such as power outages.

\textbf{Seamless indoor-outdoor model:} Basically, IPS can operate completely independent of any external position information, just by tracking the visual features with the stereo camera pair and fusing it with the built-in inertial measurement unit (IMU). But to be combined with other available data, especially the aerial images and the digital object model (DOM) provided by MACS-SaR, IPS requires an external reference to be able to output the geometric information in global coordinates.
While walking outside in areas with good GPS reception the coordinate system of IPS can be aligned with the global coordinates. But a rather long walk with very good GPS reception is necessary to reliably achieve the high alignment and precision, required for indoor walks without GPS. Especially in extended indoor walks the rotational error in the alignment propagates linearly to the distance travelled without GPS causing the main error in localization. Therefore another practical solution was developed in the context of the INGENIOUS project. 
This work builds upon our previously published approach for IPS-based thermal mapping and GNSS-independent 3D localization \cite{Schischmanow2022}, which focused primarily on infrastructure inspection scenarios. In contrast, the present study emphasizes the integration and operational deployment of the IPS system within the context of disaster response, specifically as part of the EU-funded INGENIOUS project. Here, we address new challenges related to real-time indoor-outdoor navigation for First Responders, integration with the Multi-sensor Assessment and Context-sensitive System for Search and Rescue (MACS-SaR), and field validation in safety-critical environments. While both systems share common technological components, the scope, target users, and application scenarios are fundamentally distinct.

As MACS-SaR is first flown over the disaster area, it is possible to position automatically detectable tags on random (but rather plain) places  in front of buildings IPS is supposed to enter. In the aerial images of MACS the tags can be recognized fully automatically and their positions can be triangulated using the precise position and orientation information of the airborne system. Global coordinates of the tags are then transmitted to IPS in a small text file. Before entering a building, IPS points to at least three of these Tags in front of the building to align its coordinate system with the global coordinate system. 

The necessary steps to achieve this are explained in more details in the following Section. The findings are subsequently detailed in the results Section (Section \ref{sec:results}).

\section{Materials and Methods}\label{sec:materialsandmethods}

\subsection{MACS-SaR}\label{sec:materials_macs}
\begin{figure}
	\includegraphics[width=1\linewidth]{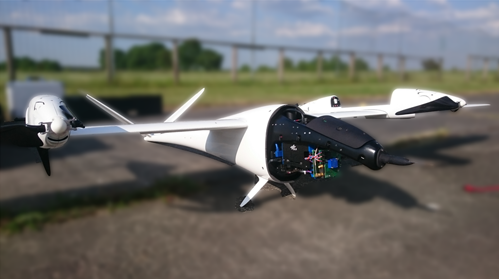}
	\caption{Prototype of the camera system mounted to carrier}
	\label{fig:macs-sar}
\end{figure}

\begin{figure}
	\includegraphics[width=1\linewidth]{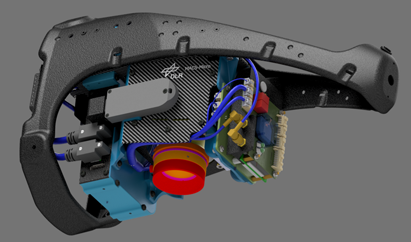}
	\caption{CAD-model of the camera system}
	\label{fig:macs-sar-head}
\end{figure}

To account for various disaster scenarios MACS-SaR is used in conjunction with a versatile carrier system coming from Quantum Systems. Due to national law policies the maximum take-off weight (MTOW) is often restricted which also leads to limitations for the payloads.
Using carriers with an MTOW less 5kg is quite popular in Germany \cite{BMVI2016} which was the starting point to develop a lightweight but metric aerial camera system \cite{dlr104190} with the intention to verify the system using traditional photogrammetric evaluation procedures \cite{dlr111102}.
This investigation confirmed that the prototype is a camera system with long term stable interior orientation.\\[1ex]
Based on this precision camera system the first prototype of the drone-based real-time mapping camera was developed in 2018 (see Figure \ref{fig:macs-sar}).
The system incorporates an industrial camera, a dual-frequency GNSS receiver featuring inertial-aided attitude processing (INS), and an embedded computer.
At the core of the camera head is a 16~MPix CCD sensor (ON Semiconductor KAI-16070 with Bayer pattern) paired with an industrial F-Mount lens (Schneider Kreuznach Xenon-Emerald 2.2/50).
To optimize image quality, the aperture is set to f4.0, and the focus is fixed to the hyperfocal distance.
Exterior orientation calculation is based on a dual-antenna GNSS receiver (Novatel OEM7720) in combination with an industrial grade MEMS-IMU (Epson G320N).
Using a dual-antenna configuration allows for true-heading determination independent of the INS.
This setup enhances orientation accuracy, particularly when movement direction and heading diverge due to cross-wind.
Depending on the flight trajectory, divergences of up to 10 degrees between movement direction and heading have been observed.
Additionally the dual-antenna system allows for very fast attitude initialization already on ground without aircraft movement.
With a baseline of approximately ~0.95~m, the two GNSS antennas are mounted at the front and tail. 
Position and attitude are continuously estimated by the GNSS receiver, which also processes the end-of-exposure signal.\\[1ex]
Thus, every image is assigned with precisely measured time, position and orientation.
Considering the interior camera orientation as long-term stable, direct geo-referencing can be applied.
Due to continuous synchronization of all subsystems, each image can be time-stamped with a precision better than 1$\mu$s.
Time synchronization, image acquisition and real-time image processing is done by the embedded computing unit.
This computer is powered by a Quad Core Processor (Intel Atom E3950) with 8~GB RAM and runs a Linux operating system.
In this configuration the system allows to capture up to 4 raw images per second which can be stored on a removable storage device.
The camera system, shown in Figure \ref{fig:macs-sar-head}, weighs 1.4~kg (including embedded PC, camera, IMU, GNSS receiver, GNSS antenna, power management, and structure) with dimensions of 10~x~14~x~20~cm$^3$.\\[1ex]

A fixed-wing drone developed by Quantum Systems (see Figure \ref{fig:macs-sar}) is used as carrier, providing a flight time of approximately 90 minutes at cruise speeds between 60~km/h and 90~km/h.
Thus, the carrier is capable to travel a distance of up to 105~km per battery charge.
It is specified with MTOW of 14~kg including a payload of up to 2~kg and has a wingspan of 3.5~m.
It can operate at wind speeds of up to 8~m/s and temperatures between 0\textdegree{}C and 35\textdegree{}C.
While its typical flight operation altitude is in the range between 100~m and 300~m above ground level, it is capable of operating at altitudes up to 3,000~m above sea level.\\[1ex]
The operational range is only limited by the maximum flight time because the autopilot systems allows fully automated flights beyond visual line of sight (BVLOS).
This requires a predefined flight plan with terrain follow mode for security reasons.
The drone is equipped with a conventional command and control link as well as an additional mobile network radio for BVLOS operation.
For safety reasons this carrier is equipped with position lights and an integrated automatic dependent surveillance broadcast (ADS-B) transceiver.

\subsection{IPS}\label{sec:materials_ips}
DLR's IPS \cite{dlr63430,dlr111874} combines stereo-based optical and inertial navigation to achieve self localization without the need of any external positioning system or other aid.
Designed to operate on foreign planets without any localization infrastructure, it proves highly valuable for localizing assets and personnel in diverse disaster environments. Especially because in the assumed scenarios a power-failure is very possible. Thus, infrastructure-based localization via pre-installed WiFi, Bluetooth beacons or other infrastructure is not reasonable. 
Furthermore global positioning with GNSS inside buildings is very unreliable and inaccurate due to heavy signal absorption and reflection.\\[1ex]

\begin{figure}
	\includegraphics[width=1\linewidth]{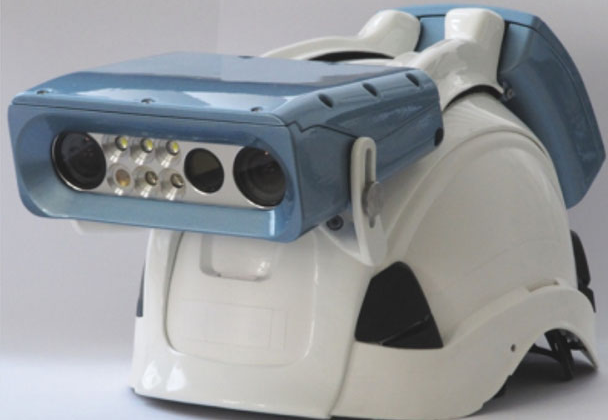}
	\caption{Helmet-version of IPS with two panchromatic cameras at the left and the right side of the frontal blue box. In between there is an auxiliary RGB-Camera with white and infrared LEDs for optional illumination. Hidden in the blue box is the IMU (front) and the synchronization electronics (back)}
	\label{fig:ips-helmet-a}
\end{figure}

\begin{figure}
	\includegraphics[width=1\linewidth]{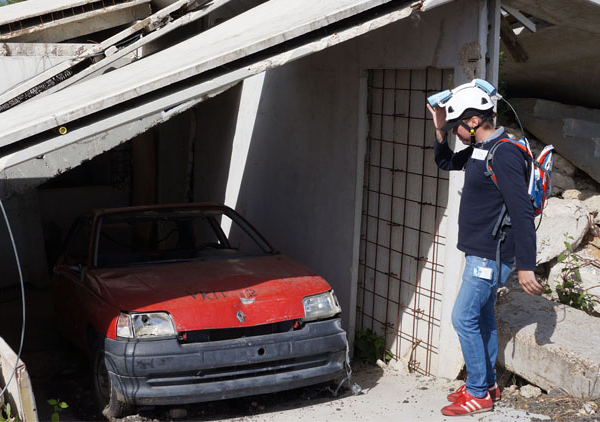}
	\caption{IPS at an exercise at the "Aire De Formation Internationale Cynotechnique Et Sauvetage Déblaiement" of CSP France}
	\label{fig:ips-helmet-b}
\end{figure}

IPS is equipped with two industrial cameras that capture image pairs over time, while a processing unit analyses the motion within these images.
Assuming that most of the objects in front of the camera remain stationary, it calculates the ego-motion from the motion in the images (optical navigation).
The motion calculation is densely coupled with the built-in inertial measurement unit (IMU) measuring the acceleration and the speed of rotation of IPS in all three dimensions.
In both the motion tracking and the filter approach, the IMU and optical navigation complement each other, resulting in accurate and robust self-localization of the IPS. \\[1ex]

The two stereo cameras used for navigation are CMOS cameras with 1280 x 1024 pixels (Model Ximea MQ013RG). 
An additional RGB camera used for inspection images is of the same type as the stereo cameras, but has a RGB pattern (Ximea MQ013CG). 
If needed, both cameras can be supported by flashing LEDs in low-light or dark scenarios. 
While a white LED flashes synchronously to exposures of the RGB camera, the navigation cameras are supported by NIR LEDs. 
This way the user is not molested by the continuous high-frequent flashes (because the human eye is not sensitive to NIR light) but power consumption and heat emission of the LEDs can be reduced to a minimum (in contrast to lightning up the LEDs continuously). 
The inertial data is measured with an IMU (ADIS16488), which is mounted rigidly on an aluminium bar together with the cameras.
All data is time-synchronized using an individually designed FPGA board with an accuracy on microsecond level.
 
In order to provide a hands-free operation for the First Responders in the project INGENIOUS the helmet version of IPS is used as a starting point (see Figure \ref{fig:ips-helmet-a}). It is extended for the integration with INGENIOUS toolset.
Within the toolset IPS has three main tasks. In real time it shall:
\begin{itemize}
	\item	keep track of its ego-position and orientation 
	\item	take user-triggered photos with the RGB camera and tag it with the current position and orientation
	\item	build up a sparse 3D model of the environment travelled through
\end{itemize}
and send all this data in world coordinates (UTM) and in real time to the Center of Operations, which then combines and displays this information for further use and distribution to other first responders.

In a post processing step a dense 3D point cloud is created. 
Therefore, the IPS stereo image pairs were used to generate sequences of high-density depth maps. 
Fusing them with the high accuracy trajectory with six degrees of freedom generated by the navigation framework of IPS, a dense 3D point cloud for the whole observed area is formed. 
The necessary frame rate of depth maps and local 3D point clouds for the subsequent aggregation of a sufficiently dense global 3D point cloud is adjusted based on the navigation solution, see \cite{Schischmanow2022}.

The IPS helmet can record RGB camera data in addition which are also synchronized with the other sensors. 
They can be mapped immediately onto local parts of the 3D point cloud in order to enrich the 3D scene with color information.

Therefore, a trifocal geometrical calibration as described in \cite{Choinowski19} is mandatory. 
A voxel-based occlusion algorithm is applied to prevent incorrect color assignment on occluded 3D points which could be caused by the side by side placement of the (left) stereo and the RGB camera. 
In a subsequent automatic filtering process, the aggregated partial point clouds are cleaned. 
Voxels are removed based on the number of 3D points found per voxel, the proportion of values with and without RGB information and their reliability. 
Cloud and voxel filtering steps are based on modules of Point Cloud Library \cite{Rusu.2011}. 
If enough computing power is available these relatively expensive calculation steps (applying image rectification, a semi-global matching algorithm and voxel filter operations) could be done online, otherwise in an immediately following post-processing step on site.

\subsection{Co-registration}\label{sec:materials_coregistration}
IPS comes along without any positioning information and therefore navigates in a local coordinate system. As world coordinates were required in the context of the INGENIOUS project and because of the aim to co-visualize dense point clouds from IPS and MACS, IPS has to be co-registered with world coordinates. In order to be independent from external positioning information, which may be unavailable or unreliable in disaster and/or mixed indoor-outdoor situations, IPS is directly co-registered with the MACS imagery.

\begin{figure*}
	\includegraphics[width=1\linewidth]{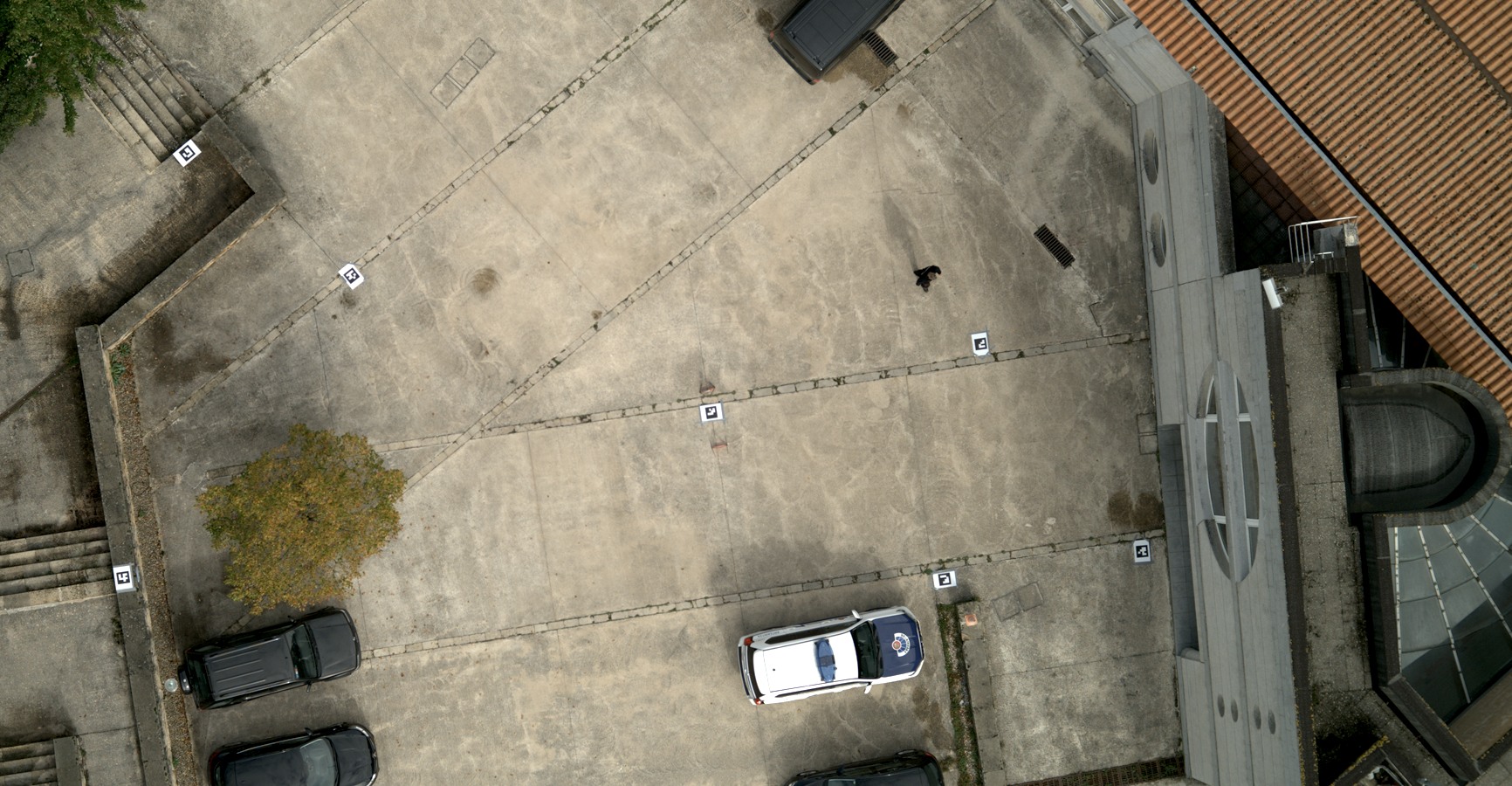}
	\caption{Part of an image from MACS showing 7 AprilTags in front of the examined building}
	\label{fig:berrozi_macs_tags}
\end{figure*}

Before MACS is started, four or more A3 sized plates with AprilTags \cite{Olson.2011} (see Figure \ref{fig:berrozi_macs_tags}) are placed in arbitrary positions that fulfil three conditions: They have to be well visible from the air, on spots where it is unlikely that they are moved unintentionally, and they must be distributed in an area with some meters of extend in two dimensions (not along a single line).

In MACS imagery the locations of the AprilTags are determined automatically with help of OpenCV-library \cite{opencv_library2000}. Having highly overlapping aerial image, all extracted image coordinates in combination with precisely measured position and orientation in world coordinate system is used to triangulate every Apriltag-centre position. These coordinates are then passed on to IPS (via ground control station).

\begin{figure}
	\includegraphics[width=1\linewidth]{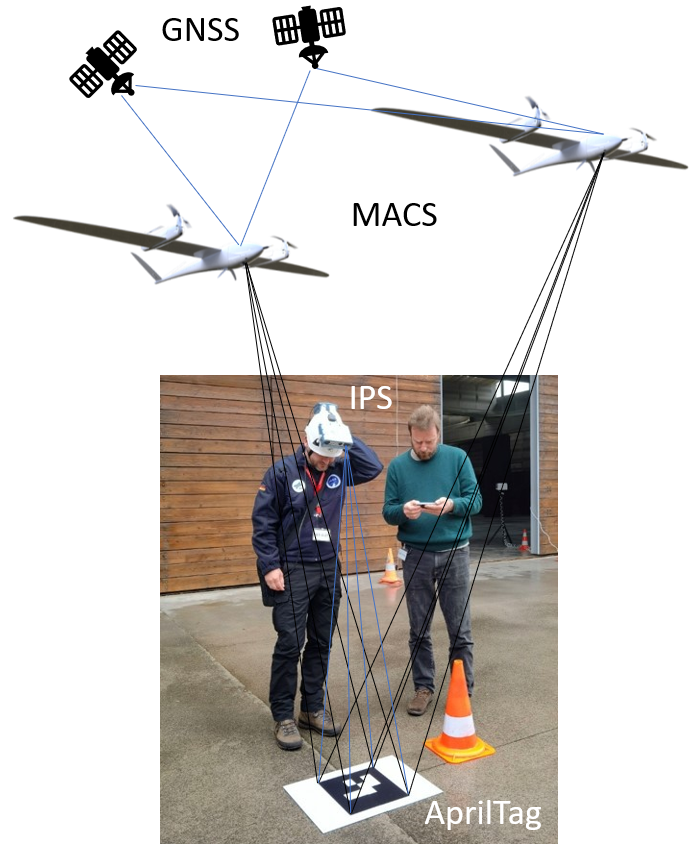}
	\caption{Visualization of the geometry underlying the co-registration process.}
	\label{fig:coregistration}
\end{figure}

IPS uses the AprilTags as global optical landmarks, to assign the local trajectory at a specific time stamp to a global reference. 
During the measurement run, one of the IPS stereo cameras takes pictures in which AprilTags can also be seen.
With the help of the OpenCV-library and the recognition and processing algorithms it contains, the AprilTags are automatically detected and converted to a 3D distance vector.
This establishes the connection between the world coordinates of the AprilTags and the local coordinate system, the moving IPS initially navigates in (See Figure \ref{fig:coregistration}).

Passing several of these optical landmarks, a 6 DoF transformation can be estimated, with which the IPS trajectory can be converted from local to global coordinates, as demonstrated in \cite{Hein19}.
After this successful switchover, IPS can calculate its position and orientation in global coordinates in the further course of the measurement run, regardless of whether it is moving inside or outside a building.

\subsection{3D Point Cloud Generation}\label{sec:materials_point_clooud_generation}

The workflow for generating digital surface models (DSM), or, in this case, a 3D building model, from remote sensing data (e.g. MACS aerial images \cite{Hein2019}) begins by refining parameters related to exterior sensor orientation (EO) and/or interior camera orientation (IO). Typically, EO parameters consist of three degrees of freedom (DOF) for translation ($X$,$Y$,$Z$) and three DOF for rotation ($\omega$,$\phi$,$\kappa$), while IO parameters commonly include the focal length ($f$), the principal point ($x_0$,$y_0$), and a set of distortion parameters (such as those in a radially symmetric distortion polynomial with coefficients  $k_0$,$k_1$,$k_2$,$\ldots$).

This orientation refinement, often referred to as bundle block adjustment (BBA), involves two main stages. The first step is to identify unique image features that serve as measurements in the image space. Next, the process minimizes the distance between the back-projected 3D estimates and the previously determined positions of corresponding features \cite{triggs1999bundle}. For each unique feature, the collection of forward-projected rays, transformed from the image coordinate system into the world coordinate system, form a bundle. Essentially, BBA solves a nonlinear least squares problem to tighten this bundle as much as mathematically possible, thereby refining both IO and EO parameters.

\begin{figure*}
	\includegraphics[width=1\linewidth]{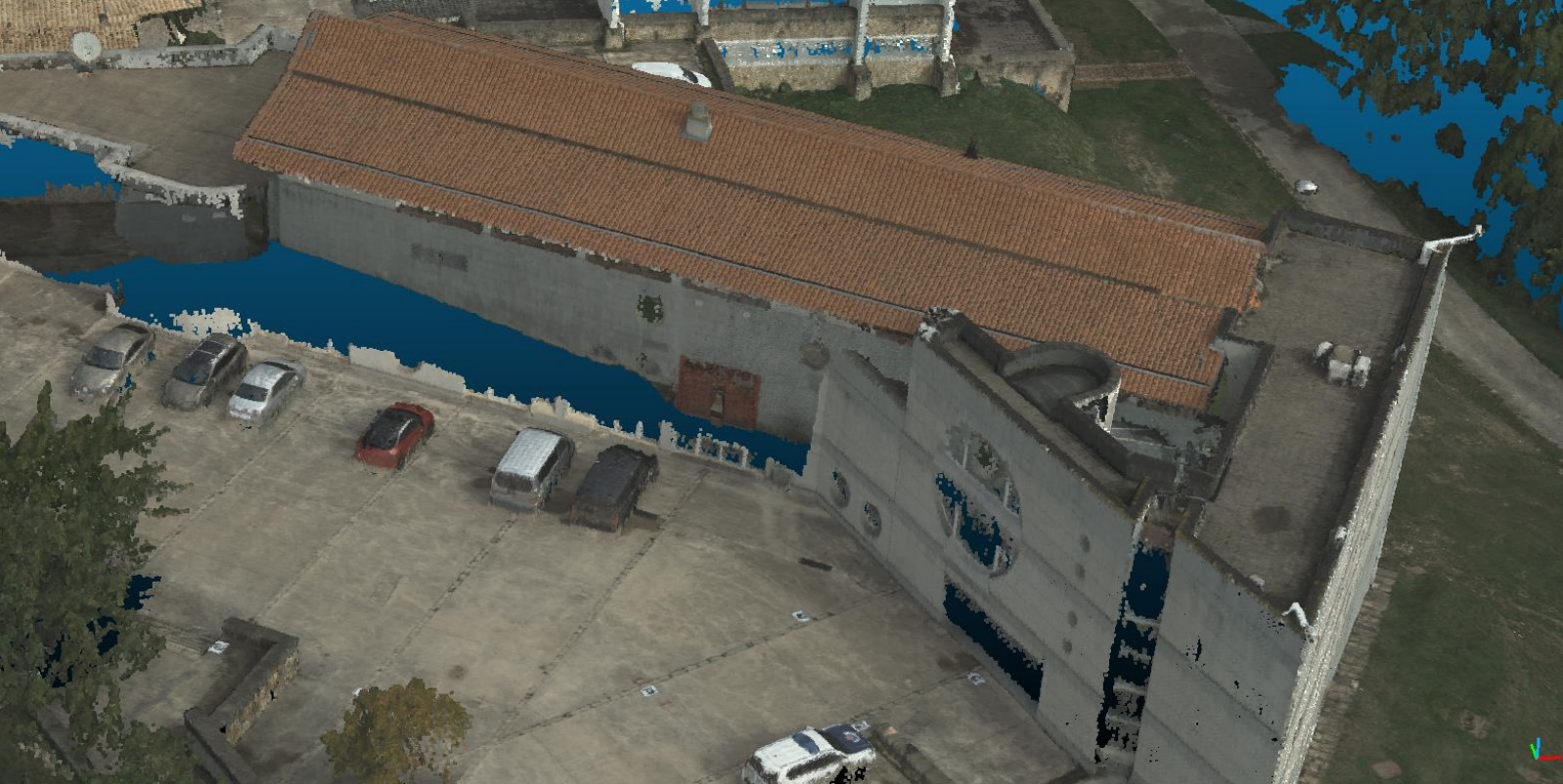}
	\caption{Point cloud in world coordinates, generated using the images of MACS.}
	\label{fig:macs_cloud}
\end{figure*}

\begin{figure}
	\includegraphics[width=1\linewidth]{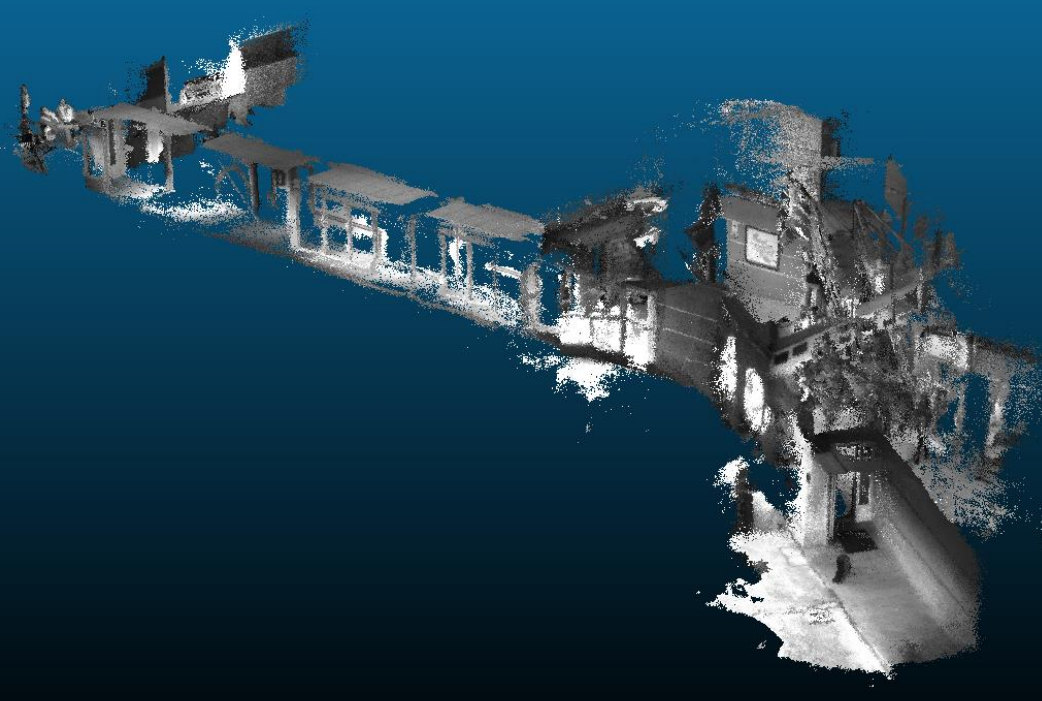}
	\caption{Point cloud in world coordinates, generated using the co-registered trajectory and stereo-images while walking through the building}
	\label{fig:ips_cloud}
\end{figure}

After obtaining all image features and their corresponding back-projected 3D coordinates, the final phase of BBA focuses on minimizing the reprojection error between observed and predicted image points. This error is quantified as the sum of squares of many nonlinear, real-valued functions. A nonlinear least squares algorithm is employed for this optimization, with the Levenberg–Marquardt method proving particularly effective \cite{Lourakis2005}. A more detailed discussion of this minimization approach is available in \cite{Triggs2003}. The refined model parameters that result from this process represent the optimized EO and/or IO values.

In photogrammetry and computer vision, these extracted and refined image features, once projected into the world coordinate frame, are typically considered a preliminary 3D reconstruction of the scene. However, in the context of this paper, the goal is to achieve a complete reconstruction with maximum detail and accuracy. Consequently, dense image matching and subsequent point cloud fusion become critical components of the 3D reconstruction workflow.

Dense image matching seeks to locate a corresponding pixel in a match image for every pixel in a designated base image, with the pixel location difference known as disparity \cite{Qian1997}. Challenges such as occlusions or non-overlapping areas between images can limit reconstruction. Several techniques exist to create a dense disparity map. One simple method examines a neighborhood around each pixel in the base image to find the best match in the corresponding match image. However, this local approach can be error-prone, especially in areas with repetitive textures or highly redundant content. Expanding this method to a global approach improves reliability, though it comes with a significant increase in computational demand \cite{Hirschmueller2012}.

To balance runtime and accuracy,  Hirschm\"uller \cite{dlr22952} introduced the Semi-Global Matching (SGM) technique. Starting with rectified images in a standard stereo setup, each pixel in the base frame is matched along its corresponding epipolar line in the match image. At every position along this line, the image is scanned along a predefined number of paths (e.g., 8, 16, etc.), extending the method into a semi-global framework. Costs based on pixel similarity, measured by metrics such as the Hamming distance or mutual information, are accumulated along each path. The final step involves minimizing an energy function for all accumulated costs $C(p,D_p)$, where $p$ represents the path and $D_p$ the associated disparity.

Within this energy minimization framework, two smoothness constraints are proposed. A constant penalty, $P_1$, is applied for small disparity changes (typically a one-pixel difference) in neighboring pixels, allowing adaptation to slanted or curved surfaces. A larger penalty, $P_2$, is imposed for more significant disparity changes to preserve discontinuities.

The SGM process ultimately produces a 3D point cloud of the observed scene, such as a building, which serves as the basis for subsequent 3D visualization. Since the IPS point cloud shares the same coordinate system, it can be directly co-visualized with MACS' point cloud (see Figure \ref{fig:IPSandMACS}).

\section{Results}\label{sec:results}

In the context of the INGENIOUS project, several locations have been used to test the developing toolkit for several of the disaster scenarios covered project. In this paper, the results of a site near Bilbao is used to visualize the steps of seamless indoor-outdoor mapping with the achieved results.

\subsection{Point Clouds}\label{sec:pointcloudgeneration}

A small part of the point cloud of the exterior, created from the MACS imagery (according to the description in Section \ref{sec:materialsandmethods}), is shown in Figure \ref{fig:macs_cloud}. This is the building IPS is supposed to enter. Therefore seven AprilTags have been positioned beforehand in front of the entrance, which can also be seen in one of the aerial images in Figure \ref{fig:berrozi_macs_tags}. Flat areas, as well as most of the facades of the building could be reconstructed, but the interior of the building cannot be sensed this way. 

But the interior of the building is explored with IPS, and a point cloud is created from the stereo images, as visualized in Figure~\ref{fig:ips_cloud}. Thanks to the co-registration, the point cloud is available in world coordinates. 

By providing accurate 3D reconstructions of the exterior and interior of buildings, the generated point clouds help First Responders navigate hazardous environments and identify potential risks or structural weaknesses. This information is crucial for planning safe rescue operations.

\subsection{Co-Visualization}\label{sec:covis}

\begin{figure*}
	\includegraphics[width=1\linewidth]{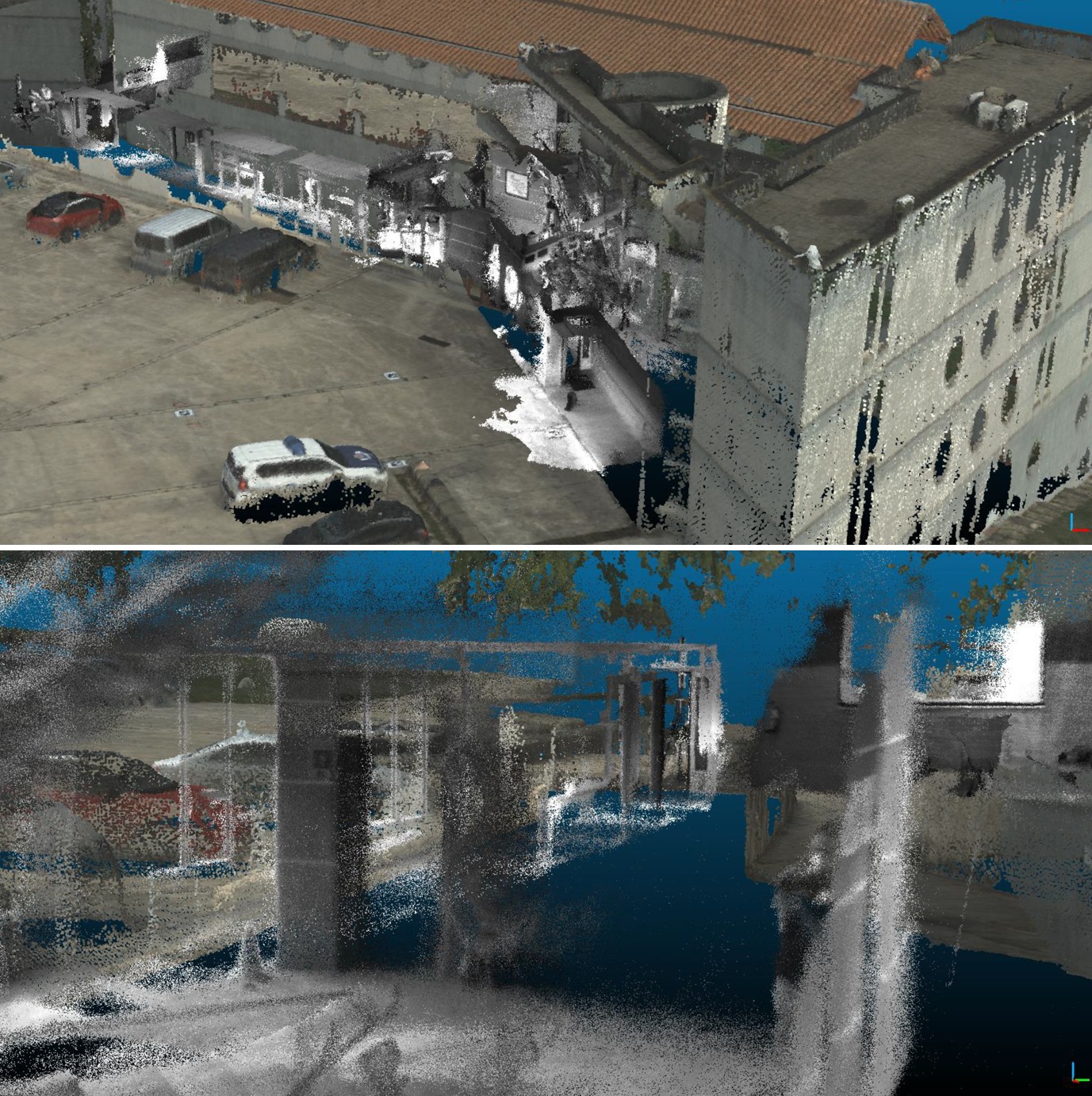}
	\caption{Co-visualization of IPS point cloud (grey) and MACS point cloud (RGB). View from outside-in (top), view from inside-out (bottom)}
	\label{fig:IPSandMACS}
\end{figure*}

As the two point clouds share the same coordinate system, they can be visualized together, as shown in Figure \ref{fig:IPSandMACS}. To allow a view inside the building from the aerial perspective, the points of the facade are removed in the upper part of the Figure. In the lower part, the view from inside the building through the window facade to the parking lot is shown. 

The co-visualization of indoor and outdoor data enables First Responders to better understand the spatial relationships between different areas of interest, facilitating coordinated actions and improving operational safety.

\subsection{Accuracy measures (absolute and relative)}\label{sec:accuracy}

Triangulation accuracy of the AprilTags has been assessed in two ways. First, all tags have also been surveyed using a GNSS-rover and the MACS triangulated coordinate have been compared to the GNSS measurements. For eight tags the average difference was $-0.23m$ in x-direction, $0.30m$ in y-direction and  $1.0m$ in z-direction.
To assess the relative net accuracy the following approach was chosen; for all possible combinations of measured world-points (position of tag 1 to position of tag 2, position of tag 1 to position of tag 3, ...) the distance was calculated. The same has been done for the triangulated points. Finally, these distances (distinguished in x,y,z) have been compared as calculated as mean of all observations delivering following results: relative difference in x-direction is $5.0cm$ in y-direction $10.6cm$ and in in z-direction $6.6cm$. 
Comparing absolute and relative accuracy reveals that even though there is a dominant absolute offset in z-direction, the relative accuracy is very good and sufficient for co-visualization. Aforementioned offset also is not of great importance because it is constant (or the same) for IPS and MACS.

\section{Conclusions and Outlook}

This work shows a method for the generation of combined indoor-outdoor models in world coordinates. Using commonly observed AprilTags the required co-registration of the aerial and indoor system proved to work reliably. 

In practice, it would also be possible to register IPS directly to the world coordinate system using the integrated GPS receiver. If the required accuracy of co-registration is low than this is also an option. But as the path IPS travels outside the building is usually much shorter than the extend of indoor space, especially the errors in heading cause a disproportionate alignment error. Especially if GNSS reception is suboptimal due to occlusions and reflections caused by the surroundings and the building itself, this can add up to several meters of misalignment of the indoor and the outdoor model.

In contrast, the presented methods with the AprilTags is independent of GNSS quality on ground and results in a clearly sufficient accuracy, at least for visualization. Especially the relative error between indoor and outdoor point cloud can be kept very low (see \ref{sec:accuracy}).

One future work will be the replacement of the AprilTags by natural landmarks. In this work, AprilTags have been chosen because they proved to be identifiable reliably from air and from the ground. This is not the case for most natural landmarks, which can look very different from different perspectives, can be repetitive (especially in man-made environments) or move significantly within short time (physically or visually, like shadows do). This leads to a much higher number of landmarks than AprilTags to reach the same reliability and accuracy of co-registration, resulting at least in more effort collecting these, at least with IPS.

Ultimately, these findings demonstrate a method that significantly enhances the ability of First Responders to operate effectively in disaster scenarios. The integration of seamless indoor-outdoor models in world coordinates provides accurate situational awareness, even in environments where traditional positioning systems fail. By improving the reliability and accuracy of co-registration, this approach ensures that responders can focus on their mission while relying on robust and precise tools for navigation and decision-making.

\section{Acknowledgements}
\ifthenelse{\boolean{anonymizeIt}}
{
Anonymized for review.
}
{
We especially thank our colleagues Patrick Irmisch $^{1,\ddagger}$, Björn Piltz $^{1,\ddagger}$, Daniel Hein $^{1,\ddagger}$, , Anko Börner $^{1,\ddagger}$ , Ralf Berger $^{1,\ddagger}$ and Jörg Brauchle $^{1,\ddagger}$, who contributed a lot to make this work possible. 
}

{
	\begin{spacing}{1.17}
		\normalsize
		\bibliography{../../common/internal-references,../../common/external-references,references} 
	\end{spacing}
}

\end{document}